# FoundBioNet: A Foundation-Based Model for IDH Genotyping of Glioma from Multi-Parametric MRI

Somayeh Farahani [1,2,3(✉)][0000-0002-5248-4110], Marjaneh Hejazi [1][0000-0002-1823-2876], Antonio Di Ieva [3][0000-0002-5341-5416], Sidong Liu [2,3][0000-0002-2371-0713]

[1] Department of Medical Physics and Biomedical Engineering, School of Medicine, Tehran University of Medical Sciences, Tehran, Iran
[2] Centre for Health Informatics, Australian Institute of Health Innovation, Macquarie University, Sydney, NSW, Australia
[3] Computational NeuroSurgery (CNS) Lab, Faculty of Medicine, Health and Human Sciences, Macquarie Medical School, Macquarie University, Sydney, NSW, Australia
somayeh.farahani@hdr.mq.edu.au

**Abstract.** Accurate, noninvasive detection of isocitrate dehydrogenase (IDH) mutation is essential for effective glioma management. Traditional methods rely on invasive tissue sampling, which may fail to capture a tumor's spatial heterogeneity. While deep learning models have shown promise in molecular profiling, their performance is often limited by scarce annotated data. In contrast, foundation deep learning models offer a more generalizable approach for glioma imaging biomarkers. We propose a Foundation-based Biomarker Network (FoundBioNet) that utilizes a SWIN-UNETR-based architecture to noninvasively predict IDH mutation status from multi-parametric MRI. Two key modules are incorporated: Tumor-Aware Feature Encoding (TAFE) for extracting multi-scale, tumor-focused features, and Cross-Modality Differential (CMD) for highlighting subtle T2–FLAIR mismatch signals associated with IDH mutation. The model was trained and validated on a diverse, multi-center cohort of 1,705 glioma patients from six public datasets. Our model achieved AUCs of 90.58% ± 1.25, 88.08% ± 3.08, 65.41% ± 3.35, and 80.31% ± 1.09 on independent test sets from EGD, TCGA, Ivy GAP, RHUH, and UPenn, consistently outperforming baseline approaches ($p \leq 0.05$). Ablation studies confirmed that both the TAFE and CMD modules are essential for improving predictive accuracy. By integrating large-scale pretraining and task-specific fine-tuning, FoundBioNet enables generalizable glioma characterization. This approach enhances diagnostic accuracy and interpretability, with the potential to enable more personalized patient care.

## 1 Introduction

Gliomas are the most common primary brain tumors in the central nervous system [1]. The 2021 WHO Classification of Tumors of the Central Nervous System emphasizes the importance of molecular profiling—particularly the determination of isocitrate dehydrogenase (IDH) mutation status—for accurate diagnosis and prognostication [2]. However, conventional methods for assessing IDH mutation rely on invasive tissue sampling, which carries risks such as bleeding and infection and may not fully capture the tumor's spatial heterogeneity [3]. This challenge highlights the need for reliable, noninvasive approaches to glioma molecular subtyping.

Magnetic resonance imaging (MRI) is a promising modality for this purpose due to its routine clinical use and its capacity to capture diverse tissue characteristics across multiple sequences [4]. For instance, IDH-mutant gliomas may exhibit well-defined margins and a characteristic T2–FLAIR mismatch sign, while IDH-wildtype tumors tend to display less distinct boundaries and heterogeneous signal patterns [5]. Despite these discernible imaging features, accurately predicting IDH mutation status from MRI remains challenging because of intratumoral heterogeneity and the subtle nature of these cues [6].

Early radiomics approaches attempted to address this challenge by extracting handcrafted features from manually delineated tumor regions [8]. Although these methods provided initial insights, their dependence on manual segmentation and feature engineering limited their reproducibility and accuracy [9]. More recently, deep learning techniques—especially convolutional neural networks (CNNs)—have been applied to directly predict molecular profiles from multi-parametric MRI [11], [13]. Yet, these approaches often struggle with limited annotated data and the complex variability inherent in glioma imaging [13].

To overcome these obstacles, foundation deep learning models have been developed to use large-scale, self-supervised pretraining to learn robust, task-agnostic representations. These models improve generalizability and performance, especially in scenarios with limited annotated data [14]. Building on this approach, we introduce a specialized framework, Foundation-based Biomarker Network (FoundBioNet), for noninvasive IDH mutation prediction. FoundBioNet is based on the SWIN-UNETR architecture from the BrainSegFounder model [15], which was pretrained using a two-phase strategy on over 42,000 brain MRI cases, including both healthy and diseased subjects, and achieved superior segmentation accuracy on benchmark datasets. Expanding on this strong model, FoundBioNet integrates two novel modules: a Tumor-Aware Feature Encoding (TAFE) module to extracts multi-scale, tumor-specific features to capture nuanced imaging patterns associated with molecular markers; and a Cross-Modality Differential (CMD) module to highlight subtle imaging cues, particularly the T2–FLAIR mismatch, to enhance the detection of IDH mutant cases.

We train and validate the FoundBioNet model on a diverse, multi-center dataset of 1,705 glioma patients, demonstrating its strong generalizability and superiority over baseline methods. Through large-scale pretraining and task-specific fine-tuning, our tumor-centric approach enhances predictive accuracy and interpretability in noninvasive glioma characterization.

# 2 Method

## 2.1 Patient Population

We collected preoperative MRI scans from 2,428 glioma patients across six public datasets: The Cancer Genome Atlas (TCGA), specifically the TCGA-LGG and TCGA-GBM collections, Ivy Glioblastoma Atlas Project (Ivy GAP), Río Hortega University Hospital Glioblastoma Dataset (RHUH), University of Pennsylvania glioblastoma dataset (UPenn), University of California San Francisco Preoperative Diffuse Glioma MRI (UCSF-PDGM), and Erasmus Glioma Database (EGD). Patients included both low-grade (grades 2 and 3) and high-grade (grade 4) CNS WHO classifications. Inclusion criteria required availability of T1-weighted (T1), postcontrast T1-weighted (T1C), T2-weighted (T2), and T2-weighted fluid-attenuated inversion recovery (FLAIR) MRI scans. After excluding cases with missing preoperative scans (228), incomplete pathology data (204), or failed preprocessing (81), 1,705 patients remained in the final cohort (354 IDH-mutant, 1,351 IDH-wildtype) (Table 1).

**Table 1.** Patient characteristics across the six datasets: TCGA (The Cancer Genome Atlas), Ivy GAP (Ivy Glioblastoma Atlas Project), RHUH-GBM (Río Hortega University Hospital Glioblastoma Dataset), UPenn-GBM (University of Pennsylvania Glioblastoma Dataset), UCSF-PDGM (University of California San Francisco Preoperative Diffuse Glioma MRI), and EGD (Erasmus Glioma Database). Class distributions for molecular and histological grades are reported as counts and percentages.

| **Datasets** | **TCGA** (n = 213) | **UCSF-PDGM** (n = 489) | **EGD** (n = 415) | **Ivy Gap** (n = 34) | **UPenn** (n = 514) | **RHUH** (n = 40) |
|---|---|---|---|---|---|---|
| **Grade** | | | | | | |
| 2 | 47 (22%) | 56 (11%) | 119 (16%) | 0 | 0 | 0 |
| 3 | 59 (28%) | 43 (9%) | 78 (11%) | 0 | 0 | 0 |
| 4 | 107 (50%) | 390 (80%) | 474 (66%) | 22 (65%) | 610 (100%) | 40 (100%) |
| Unknown | 0 | 0 | 48 (7%) | 12 (35%) | 0 | 0 |
| **IDH** | | | | | | |
| Mutated | 89 (42%) | 103 (21%) | 139 (20%) | 3 (9%) | 16 (2%) | 4 (10%) |
| Wildtype | 124 (58%) | 386 (79%) | 276 (38%) | 31 (91%) | 498 (82%) | 36 (90%) |
| Unknown | 0 | 0 | 304 (42%) | 0 | 96 (16%) | 0 |

To optimize dataset utilization, we established two experimental scenarios: (1) employing the TCGA and UCSF-PDGM cohorts for training and internal validation, with the other datasets used as independent test sets; and (2) using the EGD and UCSF-PDGM datasets for training and internal validation, with TCGA serving as an external validation set. Due to the retrospective nature of this study, imaging protocols varied across institutions. To maintain clinical heterogeneity, no cases were excluded based on acquisition parameters or image quality. For datasets with available raw data (RHUH-GBM and TCGA), we applied the Integrative Imaging Informatics for Cancer Research: Workflow Automation for Neuro-oncology (I3CR-WANO) framework to perform preprocessing [16]. Whether processed using I3CR-WANO or provided preprocessed, all datasets underwent a uniform pipeline that included registration to a common anatomical space at a voxel resolution of $1 \times 1 \times 1$ mm³, bias field correction, and skull stripping to remove nonbrain tissue. Additionally, all scans were coregistered to the T1 or T1C images, normalized using z-score standardization, and cropped to dimensions of $96 \times 96 \times 96$ voxels. We evaluated performance using accuracy (ACC), AUC, F1 score (F1), Matthews Correlation Coefficient (MCC) and area under the curve (AUC). AUC measures how well the model separates classes across thresholds, while F1 balances precision and recall. MCC is especially important for imbalanced data, as it considers all parts of the confusion matrix and provides a balanced measure ranging from –1 (worst) to +1 (best) [17].

## 2.2 FoundBioNet Model

We build on the 62M-parameter BrainSegFounder-Tiny model with a SWIN-UNETR backbone, combining a vision transformer encoder with a U-Net–style decoder [15]. Initially pretrained using a self-supervised dual-phase strategy on large-scale neuroimaging datasets, the model was fine-tuned on the BraTS 2021 challenge [18] to capture essential anatomical and disease-specific features. For the glioma study, we further refined the stage 3 pretrained BrainSegFounder weights [15] using the UCSF-PDGM dataset with a 5-fold cross-validation approach. To ensure an unbiased evaluation, cases overlapping with the BraTS 2021 training set were excluded. The refined model was then adapted specifically for predicting IDH mutation status. As shown in Figure 1, our classification framework consists of two key modules: TAFE, which extracts multi-scale, tumor-focused features from all MRI sequences, and CMD, which enhances subtle T2–FLAIR mismatch signals. These features are then fused through a Dual-Stream Fusion (DSF) module. The network is trained end-to-end with a joint loss that balances segmentation (auxiliary task) and IDH classification (main task) objectives.

**TAFE Module:** Guided by tumor segmentation, TAFE refines feature extraction within SWIN-UNETR. The decoder produces segmentation logits $\boldsymbol{S} \in \mathrm{R}^{B\times4\times D\times H\times W}$ supervised (with Dice loss) to emphasize tumor regions. Meanwhile, the encoder generates multi-scale feature maps $\{\boldsymbol{x}_i\}_{i=1}^{4}$. Global average pooling (GAP) is applied to selected stages to form feature vectors:

$$\mathrm{X}_{(i)}^{gap} = \mathrm{GAP}\ (x_i) \in \mathrm{R}^{B\times d_i}, \tag{1}$$

The default setup uses the deepest feature $\boldsymbol{x_4}$, whereas multi-scale aggregation was evaluated in the ablation study.

**CMD Module:** The T2–FLAIR mismatch sign is an imaging marker visible on standard T2 and FLAIR MRI sequences [5]. Although it is highly specific for IDH-mutant gliomas, its sensitivity is limited [19]. To capture this sign, each T2 and FLAIR input is softly gated using the tumor probability map from the segmentation branch (whole tumor vs. background). These gated volumes pass through shared 3D convolutions to produce features $\mathrm{F_{T2}}$ and $\mathrm{F_{FLAIR}}$, and their difference is amplified:

$$\mathrm{F_{diff}} = \gamma \cdot (\mathrm{F_{T2}} - \mathrm{F_{FLAIR}})\ (\gamma > 1), \tag{2}$$

Next, channel attention is then applied:

$$\mathrm{CA}\ (F_{\mathrm{diff}}) = \sigma\ (\mathrm{MLP}\ (\mathrm{GAP}\ (F_{\mathrm{diff}})) + \mathrm{MLP}\ (\mathrm{GMP}\ (F_{\mathrm{diff}}))), \tag{3}$$

Spatial attention is computed by pooling along the channel axis, concatenating, and applying a 3D convolution followed by ReLU and a sigmoid activation. The final attention map $A_{mismatch}$ is obtained by combining channel and spatial attention through element-wise multiplication. This map re-weights the original features via residual element-wise multiplication, resulting in enhanced feature maps $F'_{\mathrm{T2}}$ and $F'_{\mathrm{FLAIR}}$, which are then aggregated using adaptive adaptive global pooling:

$$C_{\mathrm{CMD}} = \mathrm{MLP}\ ([\mathrm{GAP}\ (F'_{\mathrm{T2}}),\ \mathrm{GAP}\ (F'_{\mathrm{FLAIR}})]),\ C_{\mathrm{CMD}} \in \mathrm{R}^{B\times N_{cls}} \tag{4}$$

**DSF Module:** The DSF module fuses the classification outputs from TAFE and CMD modules ($\mathrm{C_{TAFE}}$ and $\mathrm{C_{CMD}}$) by concatenation:

$$\mathrm{C_{fused}} = [\mathrm{C_{TAFE}},\ \mathrm{C_{CMD}}], \tag{5}$$

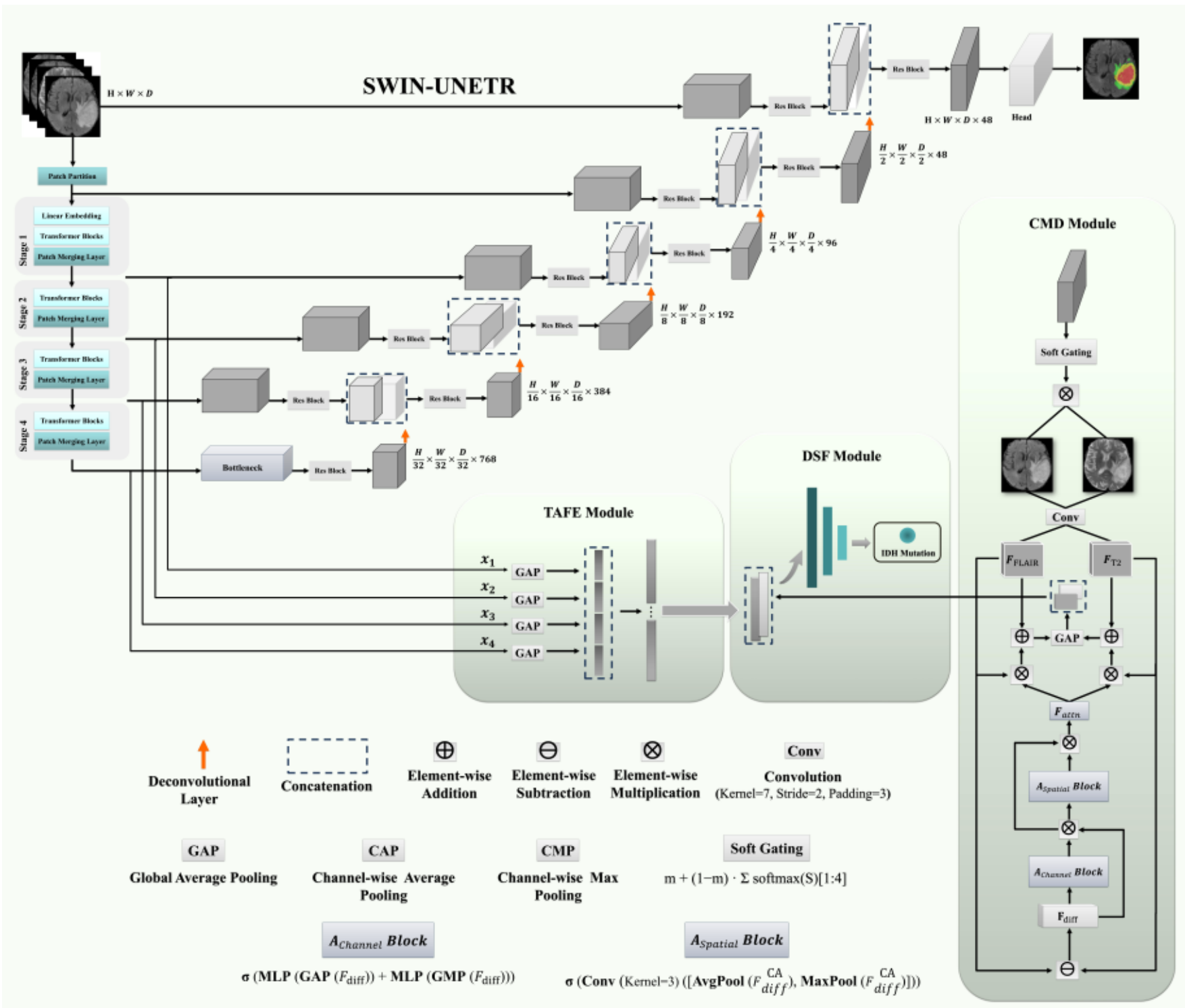

**Figure 1.** Overview of the proposed multi-task learning model. The TAFE module aggregates features from the encoder stages of SWIN-UNETR, specifically from stages $x_1$ to $x_4$. This configuration corresponds to one of the four assessed setups of the TAFE module, referred to as TAFE-4. The numbers in the subgraph indicate feature map dimensions.

A lightweight multilayer perceptron then produces the final classification logits $\mathrm{C_{final}}$. The entire network is optimized with a multi-task loss:

$$\mathrm{L_{total}} = \alpha\ \mathrm{L_{seg}}(\mathrm{S, G}) + \beta\ \mathrm{L_{cla}}\ (\mathrm{C, y}), \tag{6}$$

where ***S*** represents the segmentation logits, ***G*** the ground-truth tumor mask, ***C*** the classification logits, ***y*** the IDH mutation labels, and **α** and **β** balance the two losses.

For interpretability analysis, occlusion sensitivity maps [20] were generated using MONAI's *OcclusionSensitivity* utility by sliding a 16 × 16 × 16 voxel mask (50% overlap) across the 3D input volume and recording changes in the predicted probability for the ground-truth class. The resulting scores were Gaussian-smoothed ($\sigma = 1$), inverted, and min–max normalized to highlight regions most critical for prediction.

We used five-fold cross-validation for hyperparameter optimization, and then retrained the model and tested it on independent sets, averaging predictions from all five runs (mean ± standard deviation). Training ran on an A100 GPU (32 GB) for up to 100 epochs (batch size 2) with Adam (learning rate = 1e-4) and early stopping (patience = 5). To mitigate data imbalance and overfitting, we applied online augmentations (random flipping, rotation, intensity scaling,) and a 50% dropout. Statistical analyses were conducted in R (v4.4.1) using the rstatix package. Since the data followed a normal distribution (Shapiro-Wilk test), we applied ANOVA with post-hoc pairwise comparisons ($p < 0.05$) for the AUC metric, with the confidence interval computed using the DeLong method [21]. Our code is available on https://github.com/SomayyehF/Glioma_Biomarkers.git.

## 3 Results

### 3.1 Internal Cross-Validation and External Validation

As reported in Table 2, FoundBioNet consistently outperformed baseline models, including ViT variants—the foundation-based model for glioma analysis [22]. While models such as ResNet10, SENet101 and ViT-16 achieved competitive results during cross-validation, their performance dropped significantly on external test sets, especially when balancing MCC and F1 scores. This decline likely reflects their limited capacity to generalize to data with different imaging characteristics—a limitation that our segmentation-guided approach appears to overcome. These findings align with recent multi-task studies [27], [31], which emphasize the benefits of simultaneously analyzing tumor localization and genotype. However, our study further demonstrates that a foundation model-based approach enhances both stability and generalizability across diverse patient cohorts.

### 3.2 Ablation Experiments

Table 3 details the ablation results for the TAFE and CMD modules, both individually and combined through DSF, on the EGD test cohort. When used alone, the TAFE module achieved an AUC of 84.38%, while the CMD module yielded slightly higher performance across all metrics. Notably, combining both modules improved the results further, reaching an AUC of 90.58% with reduced variance over five runs. Although the differences between the individual modules and their fused approach were not statistically significant, the trend suggests that integrating segmentation-derived features with cross-modality differentials enhances predictive performance, consistent with a previous study [27].

To investigate TAFE more deeply, we compared it against a baseline Swin Transformer (SwinT) under four feature-aggregation depths: using only the deepest encoder feature ($x_4$, denoted "1") or progressively adding more encoder stages ($x_3 - x_4$, $x_2 - x_4$, $x_1 - x_4$). As presented in Table 4, incorporating tumor-guided information consistently improved performance across all conditions. For instance, TAFE–1 achieved an AUC of 86.62%, significantly outperforming the baseline SwinT–1 ($p < 0.05$). Increasing the feature aggregation depth without tumor guidance caused the baseline model's performance to drop sharply. In contrast, TAFE–4 maintained robust accuracy and continued to focus on clinically relevant tumor areas, as confirmed by the occlusion sensitivity maps in Figure 2. This suggests that segmentation-guided cues, regardless of depth, are more valuable than simply increasing the volume of features, aligning with recent findings [22], [23].

**Table 2.** The mean ± standard deviation of evaluation metrics for prediction models across 5-fold cross validation and external test cohorts. For a fair comparison, the ViT was trained using self-supervised pre-trained weights, while other models were trained both with and without ImageNet pre-trained weights. The best results for each model were reported. Asterisks indicate statistical significance compared to FoundBioNet, where * $p < 0.05$, ** $p < 0.001$, and *** $p < 0.0001$.

| Dataset | Model | ACC (mean±std) (%) | F1 (mean±std) (%) | MCC (mean±std) (%) | AUC (mean±std) (%) |
|---|---|---|---|---|---|
| Internal Validation | ResNet10 | 71.10 ± 2.19 | 59.77 ± 5.03 | 41.01 ± 4.04 | 82.31 ± 3.07* |
| | ResNet50 | 74.17 ± 11.14 | 67.53 ± 23.44 | 43.40 ± 10.33 | 78.60 ± 11.62 |
| | SENet101 | 74.43 ± 6.63 | 70.38 ± 14.71 | 47.64 ± 8.21 | 81.38 ± 3.64* |
| | DenseNet121 | 70.03 ± 4.71 | 62.79 ± 3.04 | 40.08 ± 3.98 | 73.65 ± 4.16* |
| | ViT-4 | 80.21 ± 5.03 | 81.19 ± 5.56 | 61.10 ± 6.02 | 88.83 ± 7.04* |
| | ViT-16 | 78.12 ± 6.19 | 76.34 ± 8.44 | 56.4 ± 6.12 | 85.83 ± 2.73 |
| | **FoundBioNet** | **90.88 ± 2.35** | **90.89 ± 2.35** | **82.06 ± 4.67** | **93.31 ± 2.46** |
| EGD | ResNet10 | 56.56 ± 1.65 | 52.20 ± 2.32 | 13.43 ± 3.36 | 56.42 ± 1.79*** |
| | ResNet50 | 55.01 ± 1.26 | 60.37 ± 3.96 | 10.78 ± 2.84 | 53.43 ± 2.75** |
| | SENet101 | 52.35 ± 4.66 | 49.10 ± 3.90 | 12.25 ± 8.10 | 60.31 ± 4.94** |
| | DenseNet121 | 62.64 ± 4.68 | 50.89 ± 6.91 | 22.21 ± 8.77 | 64.52 ± 4.58* |
| | ViT-4 | 75.01 ± 5.58 | 64.67 ± 7.04 | 46.69 ± 9.57 | 80.56 ± 5.76* |
| | ViT-16 | 72.90 ± 7.61 | 53.95 ± 15.12 | 37.45 ± 13.57 | 75.20 ± 7.60* |
| | **FoundBioNet** | **83.23 ± 1.27** | **83.70 ± 0.54** | **67.01 ± 2.10** | **90.58 ± 1.25** |
| TCGA | ResNet10 | 54.36 ± 4.59 | 43.63 ± 2.22 | 7.87 ± 2.19 | 54.98 ± 1.76*** |
| | ResNet50 | 48.77 ± 6.21 | 44.15 ± 4.51 | 4.16 ± 3.06 | 52.61 ± 2.75** |
| | SENet101 | 50.47 ± 5.47 | 56.38 ± 6.29 | 1.02 ± 11.55 | 54.12 ± 7.12* |
| | DenseNet121 | 57.35 ± 4.61 | 54.53 ± 12.38 | 14.97 ± 9.21 | 60.10 ± 4.97* |
| | ViT-4 | 68.39 ± 7.05 | 65.86 ± 11.89 | 39.25 ± 14.39 | 75.65 ± 8.44* |
| | ViT-16 | 60.79 ± 9.54 | 49.02 ± 26.95 | 23.73 ± 18.90 | 67.30 ± 10.71 |
| | **FoundBioNet** | **81.22 ± 3.60** | **79.17 ± 3.34** | **62.79 ± 6.56** | **88.08 ± 3.08** |
| Ivy GAP, RHUH | ResNet10 | 52.02 ± 1.92 | 26.05 ± 3.48 | 6.55 ± 6.08 | 38.44 ± 3.33* |
| | ResNet50 | 56.09 ± 4.94 | 34.16 ± 3.12 | 18.44 ± 13.61 | 41.01 ± 21.27 |
| | SENet101 | 53.19 ± 2.15 | 24.48 ± 8.65 | 11.30 ± 7.29 | 49.46 ± 6.49* |
| | DenseNet121 | 51.30 ± 1.79 | 11.44 ± 8.54 | 9.62 ± 9.91 | 50.45 ± 1.75* |
| | ViT-4 | 54.78 ± 3.87 | 60.15 ± 10.92 | 10.61 ± 8.79 | 51.88 ± 2.41 |
| | ViT-16 | 60.72 ± 5.35 | 53.89 ± 18.22 | 20.83 ± 13.01 | 59.15 ± 3.95 |
| | **FoundBioNet** | **66.04 ± 2.49** | **61.09 ± 2.03** | **33.51 ± 5.56** | **65.41 ± 3.35** |
| UPenn | ResNet10 | **95.38 ± 1.74** | 6.82 ± 3.03 | 5.53 ± 2.81 | 52.69 ± 3.74 |
| | ResNet50 | 89.71 ± 3.14 | 7.41 ± 2.19 | 2.32 ± 2.38 | 37.82 ± 7.56* |
| | SENet101 | 81.40 ± 8.18 | 8.09 ± 9.06 | 6.22 ± 4.78 | 63.17 ± 8.88 |
| | DenseNet121 | 90.61 ± 3.11 | 4.92 ± 3.38 | 1.65 ± 1.87 | 55.07 ± 1.98 |
| | ViT-4 | 63.43 ± 11.14 | 11.22 ± 2.21 | 13.45 ± 4.66 | 75.17 ± 7.06 |
| | ViT-16 | 73.74 ± 16.03 | 11.68 ± 4.34 | 12.76 ± 6.11 | 60.66 ± 8.43 |
| | **FoundBioNet** | 86.81 ± 4.47 | **24.11 ± 4.44** | **26.30 ± 4.14** | **80.31 ± 1.09** |

**Table 3.** Ablation studies on the three proposed modules for IDH prediction on the EGD (Erasmus Glioma Database) test dataset. The best-performing mean values are highlighted in **bold.**

| TAFE | CMD | ACC (mean±std) (%) | F1 (mean±std) (%) | MCC (mean±std) (%) | AUC (mean±std) (%) |
|---|---|---|---|---|---|
| ✓ | | 76.34 ± 8.59 | 74.53 ± 12.75 | 54.42 ± 15.70 | 84.38 ± 7.78 |
| | ✓ | 79.28 ± 4.92 | 78.50 ± 8.63 | 60.04 ± 8.08 | 85.97 ± 2.44 |
| ✓ | ✓ | **83.23 ± 1.27** | **83.70 ± 0.54** | **67.01± 2.10** | **90.58 ± 1.25** |

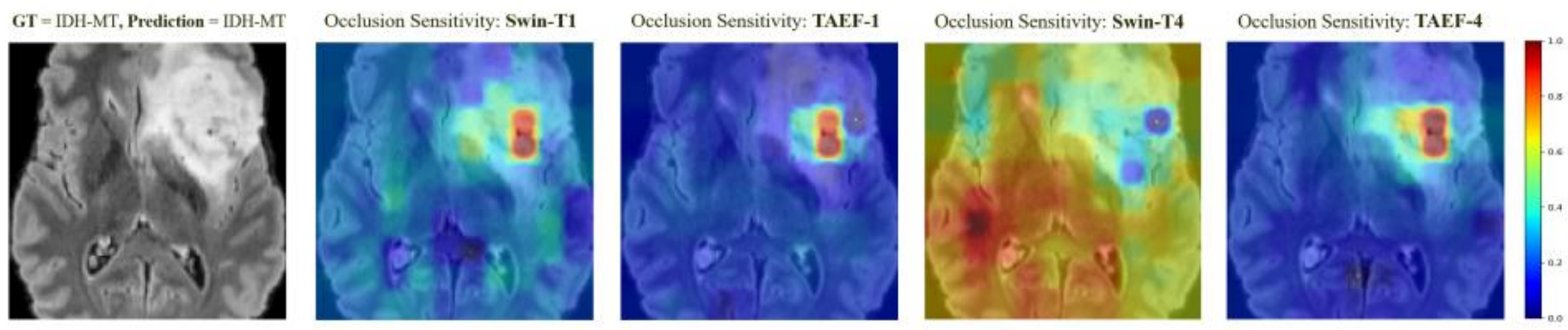

**Figure 2.** Comparison of occlusion sensitivity maps for the TAFE module (with segmentation guidance) and Swin-T (without it) under two setups: Swin-T1, TAEF-1, Swin-T4, and TAEF-4, overlaid on a FLAIR MRI slice. Swin-T1, TAEF-1, and TAEF-4 correctly predicted IDH status, while Swin-T4 misclassified it, likely due to focusing on non-tumoral regions.

**Table 4.** Comparison of segmentation guidance and feature aggregation depth on the TAFE module performance on the EGD (Erasmus Glioma Database) test dataset. The best-performing mean values are highlighted in **bold.** **'*'** denotes pairwise statistical significance between TAFE and its corresponding SwinT module, '†' indicates statistical significance compared to the SwinT-4 method, where * $p < 0.05$.

| Model Configuration | ACC (mean±std) (%) | F1 (mean±std) (%) | MCC (mean±std) (%) | AUC (mean±std) (%) |
|---|---|---|---|---|
| TAFE–1 | 74.19 ± 3.73 | 73.34 ± 7.82 | 50.86 ± 5.60 | **86.62 ± 1.67**$^{\dagger}$ |
| SwinT-1 | 70.98 ± 2.82 | 60.74 ± 2.69 | 39.25 ± 4.00 | 75.38 ± 3.87*$^{\dagger}$ |
| TAFE–2 | **76.34 ± 8.59** | **74.53 ± 12.75** | **54.42 ± 15.70** | 84.38 ± 7.78$^{\dagger}$ |
| SwinT-2 | 65.64 ± 7.89 | 49.70 ± 10.89 | 24.95 ± 15.89 | 65.66 ± 10.45 |
| TAFE–3 | 75.05 ± 3.66 | 73.56 ± 9.73 | 51.08 ± 12.95 | 80.72 ± 7.42 |
| SwinT-3 | 65.98 ± 11.57 | 55.80 ± 8.91 | 30.01 ± 18.54 | 68.73 ± 12.78 |
| TAFE–4 | 73.69 ± 7.05 | 74.49 ± 5.42 | 48.23 ± 14.03 | 80.41 ± 7.58 |
| SwinT-4 | 53.30 ± 8.22 | 50.16 ± 4.67 | 16.50 ± 10.86 | 61.86 ± 6.03 |

Although the results are promising, some limitations remain. A significant challenge is the model's reliance on accurate tumor segmentation—especially for the CMD module—to guide feature extraction, which may reduce its effectiveness when segmentation quality is poor. To mitigate this, we initialized the segmentation branch with fine-tuned weights for high-quality initialization. Rather than applying a binary crop, we employed a soft probability map with a fixed intensity floor, preserving contextual information outside the tumor. The CMD module further amplifies residual mismatch signals, enabling the model to capture even weak tumor-related cues. Global average pooling compresses feature volumes into holistic vectors. The training procedure assigns greater weight to the classification loss to prioritize IDH-relevant supervision. Additionally, early stopping, checkpointing, and fold selection are all based on IDH classification accuracy rather than segmentation Dice. For future work, segmentation reliance may be further reduced by incorporating uncertainty-gated fusion via Monte Carlo dropout [29], curriculum-based training schedules to gradually decouple segmentation guidance [30], and semi-supervised mask refinement through teacher–student consistency learning [31].

Another limitation is the model's sensitivity to class imbalance. We applied targeted online augmentations to balance the underrepresented IDH-mutant class with the wild-type class. This strategy enabled FoundBioNet to maintain strong performance on moderately imbalanced cohorts (TCGA: 42% mutant vs. 58% wild-type; EGD: 33% vs. 67%), but performance declined on highly skewed datasets such as UPenn-GBM (16 IDH-mutant vs. 498 IDH-wild-type cases). This likely stems from the limited diversity of synthetic samples, which remain within the convex hull of existing data. To address this, we plan to incorporate more advanced augmentation methods, including variants of MixUp, CutMix, and SnapMix [32], to generate more diverse and generalizable minority representations. Additionally, fixed-size cropping of input scans to manage memory constraints may occasionally exclude brain, though all cases were visually checked to ensure tumor coverage.

## 4 Conclusion

We presented FoundBioNet, a foundation-based model that noninvasively predicts IDH mutation status from multi-parametric MRI. By combining tumor-aware feature encoding with T2-FLAIR mismatch detection, FoundBioNet consistently outperforms baseline convolutional and transformer models across multiple multi-center datasets. Although further optimization is needed, our interpretable and novel approach holds significant potential for integrating advanced deep learning techniques into clinical workflows for personalized glioma management.

**Acknowledgments.** This work was partially supported by an NHMRC Ideas Grant (GNT202035).

**Disclosure of Interests.** The authors have no competing interests to declare that are relevant to the content of this article.

## References

1. Mohan, G. & Subashini, M. M. MRI based medical image analysis: Survey on brain tumor grade classification. Biomedical Signal Processing and Control 39, 139–161 (2018).
2. Louis, D. N. et al. The 2021 WHO Classification of Tumors of the Central Nervous System: a summary. Neuro-Oncology 23, 1231–1251 (2021).
3. Han, S. et al. IDH mutation in glioma: molecular mechanisms and potential therapeutic targets. Br J Cancer 122, 1580–1589 (2020).

4. Guarnera, A. et al. The Role of Advanced MRI Sequences in the Diagnosis and Follow-Up of Adult Brainstem Gliomas: A Neuroradiological Review. Tomography 9, 1526–1537 (2023).
5. Maynard, J. et al. World Health Organization Grade II/III Glioma Molecular Status: Prediction by MRI Morphologic Features and Apparent Diffusion Coefficient. Radiology 296, 111–121 (2020).
6. Park, J. E. et al. Spatiotemporal Heterogeneity in Multiparametric Physiologic MRI Is Associated with Patient Outcomes in IDH-Wildtype Glioblastoma. Clinical Cancer Research 27, 237–245 (2021).
7. Tian, Q. et al. Radiomics strategy for glioma grading using texture features from multiparametric MRI. Journal of Magnetic Resonance Imaging 48, 1518–1528 (2018).
8. Truong, N.C. et al. "Two-stage training framework using multicontrast MRI radiomics for IDH mutation status prediction in glioma." Radiology: Artificial Intelligence 6.4 (2024).
9. Hosny, A. et al. Handcrafted versus deep learning radiomics for prediction of cancer therapy response. The Lancet Digital Health 1, e106–e107 (2019).
10. Ge, C. et al. Enlarged Training Dataset by Pairwise GANs for Molecular-Based Brain Tumor Classification. IEEE Access 8, 22560–22570 (2020).
11. Li, Z. Wang, Y., Yu, J., Guo, Y. & Cao, W. Deep Learning based Radiomics (DLR) and its usage in noninvasive IDH1 prediction for low grade glioma. Sci Rep 7, 5467 (2017).
12. Chang, P. et al. Deep-Learning Convolutional Neural Networks Accurately Classify Genetic Mutations in Gliomas. Am. J. Neuroradiol. 39, 1201–1207 (2018).
13. Nalawade, S. et al. Classification of brain tumor isocitrate dehydrogenase status using MRI and deep learning. JMI 6, 046003 (2019).
14. Khan, W. et al. A Comprehensive Survey of Foundation Models in Medicine. IEEE Reviews in Biomedical Engineering 1–20 (2025).
15. Cox, J. et al. BrainSegFounder: Towards 3D foundation models for neuroimage segmentation. Medical Image Analysis 97, 103301 (2024).
16. Chakrabarty, S. et al. "Integrative imaging informatics for cancer research: Workflow automation for neuro-oncology (i3cr-wano)." JCO Clinical Cancer Informatics 7 (2023).
17. Chicco, D. et al. The Matthews correlation coefficient (MCC) is more reliable than balanced accuracy, bookmaker informedness, and markedness in two-class confusion matrix evaluation. BioData Mining 14, 13 (2021).
18. Baid, U. et al. "The rsna-asnr-miccai brats 2021 benchmark on brain tumor segmentation and radiogenomic classification." arXiv preprint arXiv:2107.02314 (2021).
19. Jain, R. et al. "Real world" use of a highly reliable imaging sign: "T2-FLAIR mismatch" for identification of IDH mutant astrocytomas. Neuro-Oncology 22, 936–943 (2020).
20. Zeiler, M. D. et al. Visualizing and Understanding Convolutional Networks. in Computer Vision – ECCV 2014 (eds. Fleet, D., Pajdla, T., Schiele, B. & Tuytelaars, T.) 818–833 (Springer International Publishing, Cham, 2014). doi:10.1007/978-3-319-10590-1_53.
21. DeLong, E. R. et al. Comparing the Areas under Two or More Correlated Receiver Operating Characteristic Curves: A Nonparametric Approach. Biometrics 44, 837–845 (1988).
22. Chen, M. et al. Medical image foundation models in assisting diagnosis of brain tumors: a pilot study. Eur Radiol 34, 6667–6679 (2024).
23. Decuyper, M. et al. Automated MRI based pipeline for segmentation and prediction of grade, IDH mutation and 1p19q co-deletion in glioma. Comput Med Imaging Graph 88, 101831 (2021).
24. van der Voort, S. R. et al. Combined molecular subtyping, grading, and segmentation of glioma using multi-task deep learning. Neuro Oncol 25, 279–289 (2023).

25. Wu, X. et al. "Biologically interpretable multi-task deep learning pipeline predicts molecular alterations, grade, and prognosis in glioma patients." NPJ Precision Oncology 8.1 (2024).
26. Chen, Q. et al. Cooperative multi-task learning and interpretable image biomarkers for glioma grading and molecular subtyping. Medical Image Analysis 101, 103435 (2025).
27. Zhang, J. et al. Multi-Level Feature Exploration and Fusion Network for Prediction of IDH Status in Gliomas From MRI. IEEE J. Biomed. Health Inform. 28, 42–53 (2024).
28. Cheng, J. et al. A Fully Automated Multimodal MRI-Based Multi-Task Learning for Glioma Segmentation and IDH Genotyping. IEEE Trans Med Imaging 41, 1520–1532 (2022).
29. Nazir, M. et al. End-to-End Multi-task Learning Architecture for Brain Tumor Analysis with Uncertainty Estimation in MRI Images. J Digit Imaging. Inform. med. 37, 2149–2172 (2024).
30. Liu, Z. et al. Style Curriculum Learning for Robust Medical Image Segmentation. Preprint at https://doi.org/10.48550/arXiv.2108.00402 (2021).
31. Felouat, S. et al. A Semi-supervised Teacher-Student Model Based on MMAN for Brain Tissue Segmentation. in Intelligent Systems and Pattern Recognition (eds. Bennour, A., Bouridane, A. & Chaari, L.) 65–75 (Springer Nature Switzerland, Cham, 2024). doi:10.1007/978-3-031-46335-8_6.
32. Qi, X. et al. MediAug: Exploring Visual Augmentation in Medical Imaging. Preprint at https://doi.org/10.48550/arXiv.2504.18983 (2025).